\documentclass[letterpaper, 10 pt, conference]{ieeeconf}
\IEEEoverridecommandlockouts                              
\pdfoutput=1                                                     

\usepackage{times}
\usepackage{graphicx}

\usepackage{amsmath,amssymb,amsfonts,amsthm,amssymb}
\usepackage{cite}
\usepackage{bbm}
\usepackage{thmtools}
\usepackage{mathtools}
\usepackage{lipsum}  
\usepackage{xspace}
\usepackage{diagbox}
\let\labelindent\relax
\usepackage{enumitem}
\usepackage{units}
\usepackage{booktabs} 
\usepackage{tabulary}

\usepackage[title]{appendix}
\usepackage{algorithm}
\usepackage{algpseudocode}
\usepackage{csquotes}
\algtext*{EndWhile}
\algtext*{EndIf}
\algtext*{EndFor}

\algblock[Input]{Input}{EndInput}
\algtext*{EndInput}
\algblock[Output]{Output}{EndOutput}
\algtext*{EndOutput}
\algblock[Variables]{Variables}{EndVariables}
\algtext*{EndVariables}
\usepackage{ifthen,version}
\usepackage{color}
\usepackage{soul}
\usepackage{moreverb,url}
\usepackage{lipsum}
\usepackage{hyperref}
\usepackage{balance}
\newtheorem{theorem}{Theorem}
\newtheorem*{theorem*}{Theorem}

\newtheorem*{lemma*}{Lemma}

\newtheorem{definition}{Definition}

\makeatletter
\@addtoreset{observation}{section}
\makeatother

\DeclareMathOperator*{\argmax}{arg\,max}

\newcommand{\argmaxprob}[1]{\argmax\limits_{#1}}

\newcommand{\expect}[2]{\mathbb{E}_{#1}\left[#2\right]}

\newcommand{\bbm}{\begin{bmatrix}}
\newcommand{\ebm}{\end{bmatrix}}

\newcommand{\set}[1]{\left\lbrace #1\right\rbrace}

\newcommand{\state}[0]{s}
\newcommand{\stateSet}[0]{\mathcal{S}}
\newcommand{\action}[0]{a}

\newcommand{\actionSet}[0]{\mathcal{A}}

\newcommand{\transFnDef}[0]{\Omega}
\newcommand{\transFn}[3]{\transFnDef{}\left(#1, #2, #3\right)}
\newcommand{\rewardFnDef}[0]{R}
\newcommand{\terminalFnDef}[0]{Term}
\newcommand{\terminalFn}[2]{\terminalFnDef{}\left(#1, #2\right)}
\newcommand{\rewardFn}[2]{\rewardFnDef{}\left(#1, #2\right)}
\newcommand{\rewardBelFn}[2]{\rewardFnDef{}_B\left(#1, #2\right)}

\newcommand{\policy}[0]{\pi}
\newcommand{\policySet}[0]{\Pi}

\newcommand{\valueFn}[2]{V^{#1}_{#2}}

\newcommand{\QFn}[2]{Q^{#1}_{#2}}

\newcommand{\valuePol}[1]{J\left(#1\right)}

\newcommand{\obs}[0]{o}
\newcommand{\obsSet}[0]{\mathcal{O}}
\newcommand{\obsFnDef}[0]{Z}
\newcommand{\obsFn}[3]{\obsFnDef{}\left(#1, #2, #3\right)}

\newcommand{\valueFnBel}[2]{\tilde{V}^{#1}_{#2}}
\newcommand{\valueFnHs}[2]{\tilde{V_{hs}}^{#1}_{#2}}
\newcommand{\QFnBel}[2]{\tilde{Q}^{#1}_{#2}}

\newcommand{\policyMDP}[0]{\pi_{\mathrm{M}}}
\newcommand{\policySetMDP}[0]{\Pi_{\mathrm{M}}}

\graphicspath{{./figs/}}

\newcommand{\xxnote}[3]{}
\ifx\hidenotes\undefined
  \usepackage{color}
  \renewcommand{\xxnote}[3]{\color{#2}{#1: #3}}
\fi

\begin{document}

\title{Hindsight is Only 50/50: Unsuitability of MDP based Approximate POMDP Solvers for Multi-resolution Information Gathering}
\author{Sankalp Arora$^{1}$, Sanjiban Choudhury$^{1}$ and Sebastian Scherer$^{1}$
\thanks{$^{1}$The Robotics Institute, Carnegie Mellon University, 5000 Forbes Avenue, Pittsburgh, PA 15213, USA 
        {\tt\small asankalp, sanjibac, basti@cmu.edu}}%
}
\date{}
\maketitle


\begin{abstract}
Partially Observable Markov Decision Processes (POMDPs) offer an elegant framework to model sequential decision making in uncertain environments. Solving POMDPs online is an active area of research and given the size of real-world problems approximate solvers are used. Recently, a few approaches have been suggested for solving POMDPs by using MDP solvers in conjunction with imitation learning. MDP based POMDP solvers work well for some cases, while catastrophically failing for others. The main failure point of such solvers is the lack of motivation for MDP solvers to gain information, since under their assumption the environment is either already known as much as it can be or the uncertainty will disappear after the next step. However for solving POMDP problems gaining information can lead to efficient solutions. In this paper we derive a set of conditions where MDP based POMDP solvers are provably sub-optimal. We then use the well-known tiger problem to demonstrate such sub-optimality. We show that multi-resolution, budgeted information gathering cannot be addressed using MDP based POMDP solvers. The contribution of the paper helps identify the properties of a POMDP problem for which the use of MDP based POMDP solvers is inappropriate, enabling better design choices.
\end{abstract}



\section{Introduction}
MDPs were developed as a part of stochastic control theory \cite{beutler1989dynamic}. The MDP formulation of problems makes underlying assumptions that the state of the environment and the agents is known, while dynamics can be stochastic with known parameters and the state is Markovian, \cite{lavalle2006planning}. MDPs are computationally relatively efficient to solve \cite{Littman:1995:CSM:2074158.2074203}. Therefore, in robotics applications, whenever it is reasonable to make the assumption that the world is completely known or the uncertainty in the world can be ignored, MDP has been a useful tool to solve sequential decision making problems \cite{choudhury2015planner, bakker2005hierarchical, trautman2010unfreezing, pivtoraiko2011kinodynamic, abbeel2007application}. However, for the problems where uncertainty cannot be ignored POMDPs were developed \cite{kaelbling1998planning}.

POMDPs are computationally intractable to solve optimally in the worst case \cite{Papadimitriou:1987:CMD:35577.35581}. Approximate POMDP solvers have been developed to solve POMDPs tractably, we refer readers to \cite{shani2013survey, ross2008online} for a comprehensive literature review. A popular approach to solving POMDPs approximately is to leverage MDP solvers and take the best expected action assuming the uncertainty about the environment will not change or it will magically disappear in the next step, FF-Re-plan \cite{little2007probabilistic}, Hindsight optimization \cite{yoon2008probabilistic} and QMDP \cite{littman1995learning} are examples of such methods. Despite strong assumptions these methods found early success as leading online POMDP solvers and have been effectively applied to problems like shared autonomy \cite{javdani2015shared}, manipulation planning \cite{koval2016pre, koval2016configuration} etc. We call such methods MDP based approximate POMDP solvers, MDP-POMDP solvers in short.

Use of MDP based approximate POMDP solvers requires computing expectation over either prior or posterior of the distribution over the state space. For large state spaces computing this expectation online can be challenging, to overcome this challenge, imitation learning techniques are used in conjunction with MDP solvers to train a policy offline such that it implicitly takes the expectation over uncertainty over state space and picks the best action online \cite{7989043, 7989112, zhang2016learning}.

MDP based approximate POMDP solvers, including the imitation learning based methods that rely on MDPs fail at taking actions that do not belong to optimal MDP policies. It is well known, actions that lead to reduction in uncertainty about the state space, while not providing rewards are an important class of actions that are often not included in optimal MDP policies \cite{yoon2008probabilistic, littman1995learning, choudhury2017data}.
The failure of agents to take such actions can lead to sub-optimal POMDP solutions, sometimes, catastrophically so. \textit{We provide a formal definition for such actions in this paper and identify with examples the conditions under which MDP based approximate POMDP solvers fail. The presence of actions identified in this paper, can be confirmed without solving full POMDP problems, and for any problem where such actions exist, the readers are advised to exercise caution against using MDP-POMDP solvers. We also show that such actions do exist for multi-resolution informative path planning problems.} The primary need to identify problems where MDP-POMDP solvers do not work is to avoid the implementation effort before realizing the unsuitability and to make better solver design choices.

Consider the famous tiger problem, \cite{kaelbling1998planning}, there is a tiger hidden behind one of the two doors in front of the agent, the agent has three possible actions. 1. Listen to determine the door behind which the tiger is hidden. 2. Open the door without the tiger for high reward 3. Open the door with the tiger for a high penalty. When the uncertainty about tiger's location is removed the optimal MDP policy is to open the door without the tiger. The action to listen for the location of tiger is never a part of the optimal MDP policy. Hence, if there is a $50\%$ chance that the tiger is behind left door and the agent is using MDP-POMDP solvers, it will never listen for location of the tiger and have only a $50/50$ chance of opening the door with a tiger in it. Whereas, if the agent would have listened for the the presence of the tiger, it could have been guaranteed to avoid the tiger. Knowing the tiger problem is unsuitable for MDP-POMDP solvers, we can avoid those methods while designing an approach to solve the problem.  

We formalize the conditions under which MDP-POMDP based solvers fail alongside the actions symptomatic of such conditions in section \ref{sec:evi}. We then provide a detailed example of the tiger problem and multi-resolution, budgeted informative path planning problem where such conditions exist and MDP-POMDP solvers can lead to unacceptably poor performance in section \ref{sec:examples}. Before going further we formally re-introduce the MDP and POMDP formulations in section \ref{sec:problem_formulation} for the sake of completeness and briefly describe how MDP solvers are used solve POMDP problems approximately in section \ref{sec:mdppomdp}.




\section{Problem Formulation}
\label{sec:problem_formulation}
\subsection{MDPs}
A markov decision process can be represented by the tuple $MDP = (\stateSet, \state_0 ,\actionSet, \transFnDef, \rewardFnDef, \terminalFnDef)$ where
\begin{itemize}
\item $\stateSet$ is a set of states.
\item $\state_o$ is the state at time $0$, initial state.
\item $\actionSet$ is a set of actions.
\item $\transFnDef$ is a set of state transition probabilities.
\item $\rewardFnDef: \stateSet \times \actionSet \to \mathbb{R}$ is the reward function.
\item $\terminalFnDef: \stateSet \times \actionSet \to \{0,1\}$ is the terminating condition.
\end{itemize}

At each time step, the agent takes an action $\action \in \actionSet$ which causes the environment to transition from state $\state$ to state $\state' \in \stateSet$ with probability $\transFn{\state}{\action}{\state'} = P(\state_{t+1} = \state' | \state_t = \state, \action_t = \action)$. The agent receives a reward $\rewardFn{\state}{\action}$. On reaching the new state $\state'$. The MDP is terminated if $\terminalFn{\state}{\action} = 1$.

$\policyMDP: \stateSet \to \actionSet$ be a policy function for the MDP that maps from a state to action.
Let the state distribution induced by a policy $\policyMDP$ after $t$ time steps, starting with state $\state_0$ be $P(\state| \policyMDP, \state_0,t)$. 
The value of a policy $\policyMDP$ is the expected cumulative reward for executing $\policyMDP$ for $T$ time steps on the induced state and history distribution
\begin{equation}
	\valuePol{\policyMDP} = \sum\limits_{t=1}^{T} \expect{\state_t \sim P(\state_t | \policyMDP,\state_0,t)}{\rewardFn{\state_t}{\policyMDP(\state_t)}}
\end{equation}

The optimal policy for the MDP maximizes the expected cumulative reward, i.e $\policyMDP^* \in \argmaxprob{\policyMDP \in \policySetMDP} \valuePol{\policyMDP}$.

The value of a state $\state$ for a given policy $\policyMDP$ is given by $\valueFn{\policyMDP}{}(\state) = \sum\limits_{i=1}^{T} \expect{\state_i \sim P(\state_i | \policyMDP,\state)} {\rewardFn{\state_i}{\policyMDP(\state_i)}}$

An MDP solver takes an MDP as an input and returns the optimal policy $\policyMDP^*$.  

\subsection{POMDPs}
In most of the real world situations the state of the environment cannot be fully known. Agents rely on observations to infer the state of the environment. Since, the state of the environment is inferred, the agent maintains a probability distribution over possible set of states that the environment can be in. Let, that distribution be $b:\stateSet \to [0,1]$. 
At each time step, the environment is in some state $\state \in \stateSet$ which cannot be directly observed. Let, the initial belief be given by $b_0(.)$. The agent takes an action $\action \in \actionSet$ which causes the environment to transition to state $\state' \in \stateSet$ with probability $\transFn{\state}{\action}{\state'}$. The agent receives a reward $\rewardFn{\state}{\action}$. On reaching the new state $\state'$, it receives an observation $\obs \in \obsSet$ according to the probability $\obsFn{\state'}{\action}{\obs} = P(\obs_{t+1} = \obs | \state_{t+1} = \state', \action_t = \action)$. 
$b_t(.)$ is the state of the belief of the agent at time then $b_{t+1}(.)$, given an action $\action_t$ and observation $\obs_{t+1}$ by 
\begin{equation}
b_{t+1}(\state') = \eta \;\obsFn{\state'}{\action}{\obs} \sum\limits_{\state \in \stateSet} \transFn{\state}{\action}{\state'} b(\state)
\end{equation}   
where $\eta$ is a normalization constant.

$\policy(b_t): B \to \actionSet \in \policySet$ be a policy function for the MDP that maps from a belief state to action.

The reward of a belief state is given by
\begin{equation}
 \rewardBelFn{b_t}{\action_t} = \sum\limits_{\state \in \stateSet} \rewardFn{\state}{\action_t}b(\state)
\end{equation}
 
The value of a policy $\policy$ is the expected cumulative reward for executing $\policy$ for $T$ timesteps on the induced belief distribution. Given a starting belief $b$, let  $P(b'| b, \policy, i)$ be the induced belief distribution after $i$ timesteps. 

\begin{equation}
	\valuePol{\policy} = \sum\limits_{t=1}^{T} \expect{b_t \sim P(b_t | \policy, t)}{\rewardBelFn{b_t}{\policy(b_t)}}
\end{equation}

The optimal policy maximizes the expected cumulative reward, i.e $\policy^* \in \argmaxprob{\policy \in \policySet} \valuePol{\policy}$.

The value of executing a policy $\policy$ from a belief $b$ is the expected cumulative reward:
\begin{equation}
\label{eq:pomdp:v}
\valueFnBel{\policy}{}(b) = \sum\limits_{i=1}^{T} 
\expect{b_i \sim P(b_i | b, \policy, i)}
{\rewardBelFn{b_i}{\policy(b_i)}}
\end{equation}

The Q-value function $\QFnBel{\policy}{}(b, \action)$ is defined as the expected sum of one-step-reward and value-to-go:
\begin{equation}
\begin{aligned}
\label{eq:pomdp:q}
\QFnBel{\policy}{}(b, \action) =& \rewardBelFn{b}{\action} + \\
                        &\expect{b' \sim P(b' | b, \action)}{\valueFnBel{\policy}{}(b')}
\end{aligned}
\end{equation}

POMDP provides an elegant framework to formalize and tackle planning problems for agents operating in partially known environments. There are two major challenges that POMDP solvers face: (1) \textit{Keeping track of evolving uncertainty about the state space over the planning horizon.} As future observations need to be accounted for, the solver needs to keep track of future beliefs that are exponential with respect to number of future observation steps. (2) \textit{Computing the expectation over the state space.} Since the state space of most of the problems worth solving is large, computing an expectation over such state space needs large computation, making it expensive to evaluate online.  
In the next section we explore how MDP-POMDP solvers overcome these challenges.
\section{Solving POMDP through MDP Solvers}
\label{sec:mdppomdp}
Approximate POMDP solvers like Hindsight Optimization \cite{yoon2008probabilistic} and QMDP \cite{littman1995learning} leverage MDP solvers and simplifying assumptions about the environment uncertainty to provide approximate solutions to POMDP. Hindsight optimization finds the optimal action assuming that uncertainty about the state cannot be changed, whereas QMDP assumes that the uncertainty will magically disappear after the next action. For both these approaches, action for a given belief $b$ is given by equation \ref{eq:hindsight:action}. The only difference is that the value function computed by using QMDP is a tighter upper-bound than Hindsight optimization for the POMDP value function.

\begin{equation}
\begin{aligned}
\label{eq:hindsight:action}
\action = \argmaxprob{\action \in \actionSet} \expect{\state \sim b(\state)} {\max_{\policy_M \in \Pi_M} \QFn{\policy_M}{}(\state, \action)}
\end{aligned}
\end{equation}

Through the simplifying assumptions about the evolution of belief state, these approaches overcome the first challenge for POMDP solvers --- keeping track of evolving uncertainty over planning horizon. However, these approaches still need the expectation over the state space to be computed.

Imitation learning based approaches \cite{7989043, 7989112, zhang2016learning} address this concern through data driven techniques. MDP solvers are used over sampled MDP problems to train a policy on the expected distribution of problems. Enabling the online policy to take the best decision in expectation if magically all the uncertainty would disappear in the next step.

Family of approximate POMDP solvers that use MDP work quite well in problems where the required changes in belief can be attained by actions that are rewarding as well. A well-known problem with MDP solvers is that they do not take actions that don't belong to optimal MDP policies. Hence, neither do the imitation learning methods.

A class of such actions is informative actions. MDP solvers have no motivation to gain information, since under their assumption, either the environment is already known as much as it can be or the uncertainty will disappear after the next step but for POMDP problems gaining information can be useful, \cite{littman1995learning}.

In the next section we show that gaining information through observations can only lead to a belief state with higher value, and discuss a class of information gaining actions that will not be taken by an MDP solver leading to sub-optimal solutions by MDP-POMDP solvers.
\section{Expected Value of Information}\label{sec:evi}
Information is gained through observing the environment.
Let a POMDP be given by $POMDP = (\stateSet, b_0 ,\actionSet, \transFnDef, \rewardFnDef, \obsFnDef, \terminalFnDef)$. 
Let $\obs \in \obsSet$ be observed and the belief changes from $b$ to $b^\obs$.
The $Q-value$ of the $b^\obs$ is given by equation \ref{eq:q_value_bo}.
\begin{equation}
\begin{aligned}
\label{eq:q_value_bo}
\valueFnBel{\policy^*}{}(b^\obs) = \QFnBel{\policy^*}{}(b^\obs, \action) =& \max_{\action \in \actionSet} \bigg[ \rewardBelFn{b^\obs}{\action} + \\
                        &\expect{b' \sim P(b' | b^\obs, \action)}{\valueFnBel{\policy^*}{}(b')} \bigg]
\end{aligned}
\end{equation}
Let's say the probability of observing observation $\obs$ given a belief $b$ is $P(\obs|b)$.
\begin{definition}
\label{def:evi}
Expected value of information is given by the difference between the expected of value of belief state reached after making an observation and the optimal value of the belief state before the observation, equation \ref{eq:veq}.
\begin{equation}
\begin{aligned}
\label{eq:veq}
\mathcal{EVI}_{\obs}(b) = \expect{\obs \sim P(\obs|b)}{\max_{\action \in \actionSet}\QFnBel{\policy^*}{}(b^\obs, \action)} - \valueFnBel{\policy^*}{}(b)
\end{aligned}
\end{equation}
\end{definition}

\begin{theorem}
\label{thm:evi}
Expected value of information is greater than or equal to 0, for all observation and belief.
\end{theorem}
\begin{proof}
If we ignore the observation and take the same action that maximizes the Q-value of $b$, instead of $b^{\obs}$, then the observation does not affect the actions, and the expected value of the belief state. Lets say, the optimal action for $b$ is given by , $\action^*_b = \argmaxprob{\action \in \actionSet} \QFnBel{\policy^*}{}(b, \action)$.  
We replace $\max_{\action \in \actionSet}$ with $\action^*_b$ in equation \ref{eq:veq}, refer to equation \ref{eq:net_zero0}.
\begin{equation}
\begin{aligned}
\label{eq:net_zero0}
\expect{\obs \sim P(\obs|b)}{\QFnBel{\policy^*}{}(b^\obs, \action^*_b)} - \valueFnBel{\policy^*}{}(b)
\end{aligned}
\end{equation}
Expanding equation \ref{eq:net_zero0}.
\begin{equation}
\begin{aligned}
\label{eq:net_zero1}
\sum\limits_{\obs \in \obsSet}\sum\limits_{\state \in \stateSet} \rewardFn{\state}{\action^*_b}P(\state|b,\obs)P(\obs|b) + \\ 
\sum\limits_{\obs \in \obsSet}\sum\limits_{b' \in B} \valueFnBel{\policy^*}{}(b')P(b'|b,\obs,\action^*_b)P(\obs|b) - {\valueFnBel{\policy^*}{}(b)} = 0  
\end{aligned}
\end{equation}
Marginalizing over $\obs$.
\begin{equation}
\begin{aligned}
\label{eq:net_zero2}
\sum\limits_{\state \in \stateSet} \rewardFn{\state}{\action^*_b}P(\state|b) + \\ 
\sum\limits_{b' \in B} \valueFnBel{\policy^*}{}(b')P(b'|b,\action^*_b) - {\valueFnBel{\policy^*}{}(b)} \\
{\valueFnBel{\policy^*}{}(b)} - {\valueFnBel{\policy^*}{}(b)} = 0  
\end{aligned}
\end{equation}
Since we replaced a max function with a fixed action, we can infer that the minimum value of $\mathcal{EVI}_{\obs}(b)$ is $0$.
\end{proof}

Unfortunately, optimal MDP solvers do not take informative actions if the actions do not provide with reward or access to more rewarding states. 

\begin{definition}
\label{def:ai}
Informative actions $(a_I)$ are actions that are not a part of the optimal MDP policy for any state or time, but lead to observations. Formally, they are defined by following set of conditions: 
\begin{itemize}
\item $a_I \neq \policy^*(\state) \forall \state \in \stateSet$.
\item $P(b'|b,a,o) = P(b'|b,o)P(o|b)$, where $P(b^{\obs}|b,o)=1$ and $P(b'|b,o)=0 \forall b' \neq b^{\obs}$. 
\end{itemize}
\end{definition}

If informative actions lead to more valuable information that they cost, then MDP-POMDP solvers are provably sub-optimal.

\begin{theorem}
\label{thm:subopt} 
If there exists an informative action $a_I \in \actionSet$ in POMDP,  such that the expected value of information attained by $a_I$ is $\geq -R_b(b,a_I)$, then family of MDP-POMDP solvers will be sub-optimal by atleast $\mathcal{EVI}_{\obs}(b) + \rewardBelFn{b}{\action_I}$.
\end{theorem}
\begin{proof}
The difference between the Q-value of $a_I$ and Q-value of $\action^*_b$ is given by equation ~\ref{eq:sub_opt0}.
\begin{equation}
\begin{aligned}
\label{eq:sub_opt0}
\QFnBel{\policy^*}{}(b, \action_I) - \QFnBel{\policy^*}{}(b, \action^*_b)
\end{aligned}
\end{equation}
\begin{equation}
\begin{aligned}
\label{eq:sub_opt1}
 = \rewardBelFn{b}{\action_I} + \expect{b' \sim P(b' | b, \action)}{\valueFnBel{\policy^*}{}(b')} - \valueFnBel{\policy^*}{}(b)
\end{aligned}
\end{equation}
Since an observation is observed immediately after taking action $\action_I$ and definition of $a_I$, equation \ref{eq:sub_opt1} reduces to equation \ref{eq:sub_opt2}.
\begin{equation}
\begin{aligned}
\label{eq:sub_opt2}
\rewardBelFn{b}{\action_I} + \expect{\obs \sim P(\obs|b)}{\max_{\action \in \actionSet}\QFnBel{\policy^*}{}(b^\obs, \action)} - \valueFnBel{\policy^*}{}(b)
\end{aligned}
\end{equation}
\begin{equation}
\begin{aligned}
\label{eq:sub_opt2}
\rewardBelFn{b}{\action_I} + \mathcal{EVI}_{\obs}(b)
\end{aligned}
\end{equation}
\end{proof}

In the next section we show two examples, where MDP-POMDP solvers catastrophically fail on account of their inability to take informative actions, Multi-resolution informative path planning being one of them. We first start with a detailed tiger problem.

\section{Examples}
\label{sec:examples}
\subsection{Tiger Problem}
In the canonical tiger problem there are two doors and a tiger is hidden behind one of them. There are two possible states, either the tiger is hidden behind the left door $\mathcal{T}_L$ or the right door $\mathcal{T}_R$, $\stateSet = \set{\mathcal{T}_L, \mathcal{T}_R}$. 
The action set is given by, $\actionSet =\set{d_L, d_R,l}$, where $d_L$ denotes opening the door at left, $d_R$ denotes opening the door at right and $l$ denotes the action of listening to determine the location of the tiger.
Action $l$ can return observations, $\obsSet = \set{o_L, o_R}$, where $o_R$ denotes a tiger is heard behind the right door and $o_L$ denotes a tiger is heard behind the left door. Let the probability of hearing a tiger behind a door $X \in {L,R}$ if there is a tiger behind door $X$ is given by $P(o_X|\mathcal{T}_X) = 1$.

The reward of the action state pairs is given by table \ref{tab:tiger_reward}.
\begin{center}
 \begin{tabular}{c c c}
   & $\state = \mathcal{T}_L$ & $\state = \mathcal{T}_R$ \\ [0.5ex] 
 $\action = d_L$ & $\rewardFnDef(\mathcal{T}_L,d_L) = 0$ & $\rewardFnDef(\mathcal{T}_R,d_L) = 100$ \\ 
 $\action = d_R$ & $\rewardFnDef(\mathcal{T}_L,d_R) = 100$ & $\rewardFnDef(\mathcal{T}_R,d_R) = 0$ \\
 $\action = l$ & $\rewardFnDef(\mathcal{T}_L,l) = -1$ & $\rewardFnDef(\mathcal{T}_R,l) = -1$ \\
\label{tab:tiger_reward} 
\end{tabular}
\end{center}

Let the initial belief be $b_0(\mathcal{T}_L) = P(\mathcal{T}_L) = 1-P(\mathcal{T}_R) = 0.5$.
If the uncertainty is removed from this problem, the optimal MDP solver will suggest the action to open the door with no tiger behind it, $\action = d_L$ or $\action = d_R$. And since the action $l$ is neither rewarding or does not lead to a rewarding state, it will never be taken by the MDP solver. Approaches like hindsight optimization and \textit{Q-MDP} use MDP solvers to compute the next action. Both these approaches find the optimal action assuming that uncertainty about the state cannot be changed. The action is given by equation \ref{eq:hindsight:action}.

\begin{equation}
\begin{aligned}
\label{eq:hindsight:action}
\action = \argmaxprob{\action \in \actionSet} \expect{\state \sim b(\state)} {\max_{\policy_M \in \Pi_M} \QFn{\policy_M}{}(\state, \action)}
\end{aligned}
\end{equation}

For the tiger example with $b_0 = 0.5$, equation returns actions $\mathcal{T}_L$ or $\mathcal{T}_R$ with the hindsight upper-bound value $\valueFnHs{\policy^*_M}{}(b_0)=100$, whereas the actual value of action $\mathcal{T}_L$ or $\mathcal{T}_R$ is $50$. However, if the action $l$ is taken, the state of the environment is revealed leading to a Q-value of action $l$, $\QFnBel{\policy*}{}(b,l)  = 99$. Therefore, for this problem, Hindsight is only $50/50$ as it takes best expected action in hindsight and has no implicit motivation to gain information.
In this problem all the MDP-POMDP solvers discussed in section \ref{sec:mdppomdp}, will face the same problem. Making them unfit to solve the famous tiger problem. In the next section we demonstrate that MDP-POMDP solvers are unfit to solve multi-resolution, budgeted information gathering through a scaled down version of the problem.
\subsection{Multi-resolution, Budgeted Information Gathering}
Unmanned Aerial Vehicles (UAVs) have the capability to gain height and gather low resolution information at a large scale and the agility to zoom down on relevant regions of the information to gain high resolution information. This ability makes them suitable for locating objects in the environment and gain information about those objects. The UAV only receives a reward if the high resolution information about the object of interest is collected. Since, the action of gaining height and gaining low resolution information only results in reduction of uncertainty and no actual reward, we show it is an informative action and MDP-POMDP solvers are unable to exploit the ability of UAVs to gather information at multiple resolution.
\begin{figure*}[t!]
\includegraphics[width=\textwidth]{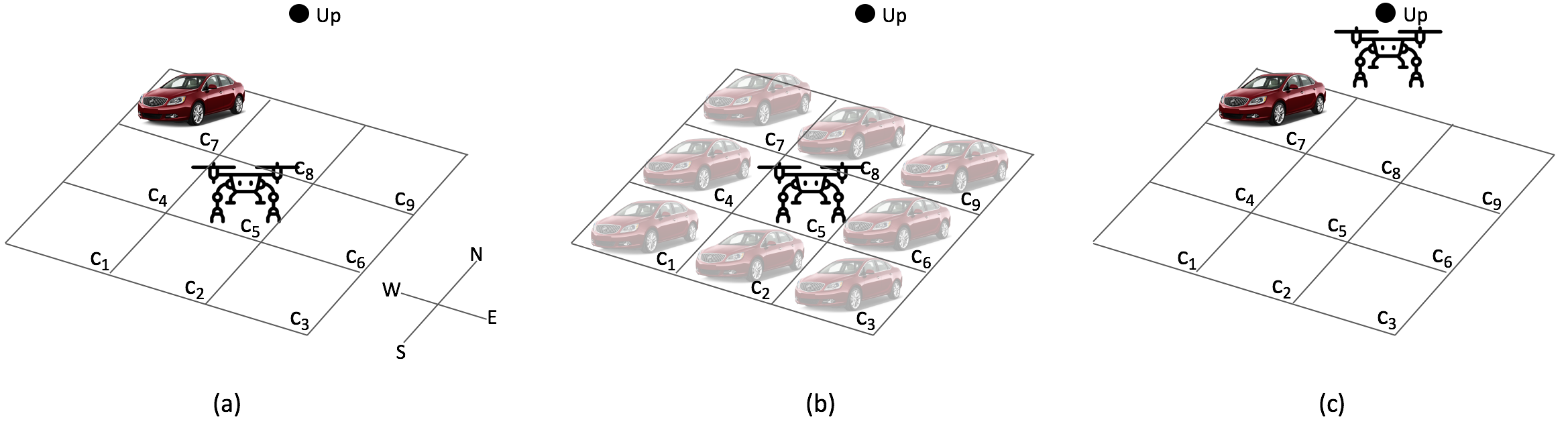}
\caption{(a) The UAV gets a high reward if it visits the same cell as the car to gather high resolution information of the car (lets say its numberplate.). Since the UAV knows where the car is, it does not need to take action $up$ and can directly go to the cell with the car. (b) In the POMDP version of the problem the location of the car is unknown, hence there is only a small chance that the next cell the UAV goes to has a car in it, leading to a small expected reward. (c) But if UAV gains height, the uncertainty about the location of the car is removed. Although there is no reward from this action but the removal uncertainty leads to guarantee of gaining high reward by visiting the cell in which the car exists. Since, MDP-POMDP solvers cannot take informative action $up$, they are sub-optimal for such data gathering problems.}
\label{fig:exploration_toy_pomdp}
\end{figure*}

Let us assume the UAV operates in a grid with cells $C = \{ c_1,c_2,c_3,c_4,c_5,c_6,c_7,c_8,c_9 \}$, with an object of interest, a car in cell $c_r$, where $r \in [1,9]$, see figure \ref{fig:exploration_toy_pomdp}. The UAV starts at time $t=0$ from cell $c_5$. At time $t$, the state of the environment is given by the history of cells visited by the UAV and the location of the car $c_r$, $\state_t = <c_r,c_{i_1},c_{i_2}, c_{i_3}, ..., c_{i_t}>$, where $c_{i_{1:t}} \in C$. There are five possible actions that the UAV can execute in this grid world, $\actionSet = \{north, east, south, west, up \}$. Action $\action = north$ moves the UAV to the north of the cell it currently occupies if possible, otherwise it keeps the UAV in the same cell. Similarly, $east, south, west$ actions move the UAV to the east, south or west of cell it currently occupies respectively. Action $up$ results in the UAV gaining height to inspect the presence of cars in all the cells and returning to the cell it executed $up$ in. Let, $T: \stateSet X \actionSet \to C$ be the deterministic transition function that takes the state of the UAV and an action as input and returns the the cell in which the UAV will be after the action is executed.

In the MDP definition of the problem, location of the car is known. The UAV gets a high reward for going to the same cell as the car, all actions cost $-1$ and action $up$ costs relatively higher, $-2$. The reward function is given by equation \ref{eq:exploration_reward}.
\begin{equation}
\label{eq:exploration_reward}
R(\state,\action) = \left\{ \,
\begin{IEEEeqnarraybox}[][c]{l?s}
\IEEEstrut
100 & if $T(s,a) = c_r \land \nexists c_{i_j} \in \stateSet:  c_{i_j}=c_r$, \\
-2 & if $a = up$, \\
-1 & otherwise.
\IEEEstrut
\end{IEEEeqnarraybox}
\right.
\end{equation}
Terminal state is reached when either cell $c_5$ is visited twice or total cost of taking actions (excluding reward of visiting the same cell as the car) exceeds five.
 
Since the location of the car is known, an optimal MDP solver will directly approach the car taking actions either $east, west, north$ or $south$ and return to the terminal cell $c_5$, the value of the starting state, $\state = <c_r,c_5>$ is given by equation \ref{eq:car_exploration_mdp_value}.
\begin{equation}
\label{eq:car_exploration_mdp_value}
\valueFn{\policyMDP^*}{}(\state =c_5) = \left\{ \,
\begin{IEEEeqnarraybox}[][c]{l?s}
\IEEEstrut
99 & if $c_r \in \{c_2,c_4,c_6,c_8\}$, \\
97 & if $c_r \in \{c_1,c_3,c_7,c_9\}$.
\IEEEstrut
\end{IEEEeqnarraybox}
\right.
\end{equation}
Action $up$ does not offer any reward and increases the cost of all paths that leads to any rewarding state. Therefore, an MDP solver will never take the action $up$. Action $up$ here is an informative action according definition \ref{def:ai}. Presence of an informative action cautions us against using MDP based Approximate POMDP solvers for this problem.

In the POMDP version of the problem, the UAV is uncertain about the location of the car. Assuming ties between actions are broken randomly, and the belief about the location of the car is uniformly distributed, the value of initial belief state $b^{uniform}_0$ given hindsight optimization is used at every step can be computed using equation \ref{eq:pomdp:v} and is given by equation \ref{eq:car_exploration_hindsight}.
\begin{equation}
\label{eq:car_exploration_hindsight}
\valueFnBel{\policy_h}{}(b^{uniform}_0) = 34.125
\end{equation}
where, $\policy_h$ is the policy attained by iteratively using hindsight optimization.
 
The value of information about the car is given by the difference between the expected value if the location of the car is known, equation \ref{eq:car_exploration_mdp_value}, the expected value if it is unknown, equation \ref{eq:car_exploration_hindsight}, $63.875$

The optimal POMDP policy will use the action $up$ to reduce the uncertainty and guarantee that it will find the car in the grid, leading to a value of $96$. As per theorem ~\ref{thm:subopt} the MDP based solvers are sub-optimal by the sum of the expected value of information and the reward achieved by the informative action, $61.875$.

As is evident, MDP-POMDP solvers perform sub-optimally at the budgeted, multi-resolution,information gathering task and fail to leverage the agility of UAVs, making them unsuitable for such applications. In the next section we discuss a few potential methods to alleviate this problem.
\section{Discussion and Future Work}
\label{sec:discussion}
Existence of informative actions, section \ref{sec:evi}, are symptomatic of unsuitability of MDP-POMDP solvers for solving a POMDP. Identification of such actions without the need to solve the full POMDP, can help avoid implementation effort to discover the unsuitability of MDP-POMDP solvers and make better design choices. Both the famous tiger problem and the multi-resolution information gathering problems are unsuitable to solve via MDP-POMDP solvers.

However, if MDP solvers itself operate in information spaces and the reward function is shaped to encourage gathering of information \cite{arora2017randomized, arora2015pasp, arora2015emergency, chen2016pomdp,7991510,
bircher_alexis_schwesinger_omari_burri_siegwart_2017,8206030,costante2016perception}, MDP-POMDP solvers can still effectively address problems with informative actions \cite{choudhury2017data}. We propose to implement such a pipeline and demonstrate its salient and failure points as compared to vanilla MDP-POMDP solvers in the future.
\balance
\bibliographystyle{IEEEtran}
\bibliography{all}


\end{document}